\pdfoutput=1

\documentclass[11pt]{article}

\usepackage{ACL2023}

\usepackage{times}
\usepackage{latexsym}

\usepackage[T1]{fontenc}

\usepackage[utf8]{inputenc}

\usepackage{microtype}

\usepackage{inconsolata}

\usepackage{amsmath,amsfonts,bm}

\def\eqref#1{equation~\ref{#1}}

\def\1{\bm{1}}

\def\rh{{\textnormal{h}}}

\def\ry{{\textnormal{y}}}

\def\vf{{\bm{f}}}

\def\vy{{\bm{y}}}

\def\evf{{f}}

\def\evy{{y}}

\def\mD{{\bm{D}}}

\def\mI{{\bm{I}}}

\def\mL{{\bm{L}}}

\def\mT{{\bm{T}}}

\def\mW{{\bm{W}}}

\DeclareMathAlphabet{\mathsfit}{\encodingdefault}{\sfdefault}{m}{sl}
\SetMathAlphabet{\mathsfit}{bold}{\encodingdefault}{\sfdefault}{bx}{n}

\def\gG{{\mathcal{G}}}

\def\sH{{\mathbb{H}}}

\def\sS{{\mathbb{S}}}

\def\sX{{\mathbb{X}}}
\def\sY{{\mathbb{Y}}}

\def\emW{{W}}

\newcommand{\E}{\mathbb{E}}

\newcommand{\R}{\mathbb{R}}

\usepackage{hyperref}
\usepackage{url}
\usepackage{amsmath}
\usepackage[ruled,vlined]{algorithm2e}
\SetKwRepeat{Do}{Do}{While}%
\usepackage{enumitem}
\setlist[itemize]{itemsep=2pt, parsep=2pt, topsep=0pt, partopsep=0pt}
\usepackage{CJKutf8}
\usepackage{graphicx}
\usepackage{booktabs}
\usepackage{array}
\usepackage{graphicx}
\usepackage{listings}
\usepackage{tabularx}  
\usepackage{tablefootnote}  
\usepackage{caption}  
\usepackage{subcaption}
\usepackage{xcolor}
\definecolor{codegreen}{rgb}{0,0.6,0}
\definecolor{codegray}{rgb}{0.5,0.5,0.5}
\definecolor{codepurple}{rgb}{0.58,0,0.82}
\definecolor{backcolour}{rgb}{0.95,0.95,0.92}
\lstdefinestyle{mystyle}{
    backgroundcolor=\color{backcolour},   
    commentstyle=\color{codegreen},
    keywordstyle=\color{magenta},
    numberstyle=\tiny\color{codegray},
    stringstyle=\color{codepurple},
    basicstyle=\ttfamily\footnotesize,
    breakatwhitespace=false,         
    breaklines=true,                 
    captionpos=b,                    
    keepspaces=true,                 
    numbers=left,                    
    numbersep=5pt,                  
    showspaces=false,                
    showstringspaces=false,
    showtabs=false,                  
    tabsize=2,
    inputencoding=utf8,
    escapeinside={(*@}{@*)}, 
}
\usepackage{multirow}  
\usepackage{tcolorbox}

\title{Enhancing Large Language Models in Coding\\ Through Multi-Perspective Self-Consistency}

\author{Baizhou Huang \textsuperscript{\rm 1,2,}\thanks{\hspace{4pt}Work done during internship at Microsoft.}$\,$
Shuai Lu \textsuperscript{\rm 3,}\thanks{\hspace{4pt}Corresponding author.}$\,$
Weizhu Chen \textsuperscript{\rm 3}$\,$
Xiaojun Wan \textsuperscript{\rm 1,2}$\,$
Nan Duan \textsuperscript{\rm 3}
\\
\textsuperscript{\rm 1}Wangxuan Institute of Computer Technology, Peking University\\
\textsuperscript{\rm 2}State Key Laboratory of Media Convergence Production Technology and Systems\\
\textsuperscript{\rm 3}Microsoft Research Asia\\
\texttt{\{hbz19,wanxiaojun\}@pku.edu.cn}, \texttt{\{shuailu,wzchen,nanduan\}@microsoft.com}
}

\begin{document}
\maketitle
\begin{abstract}
Large language models (LLMs) have exhibited remarkable ability in code generation.
However, generating the correct solution in a single attempt still remains a challenge. 
Prior works utilize \textit{verification properties} in software engineering to verify and re-rank solutions in a majority voting manner. But the assumption behind them that generated 
 verification properties have better qualities than solutions may not always hold.
In this paper, we treat them equally as different \textit{perspectives} of LLMs' reasoning 
processes. We propose the \textbf{Multi-Perspective Self-Consistency (MPSC)} framework incorporating both inter- and intra-consistency across outputs from multiple perspectives. Specifically, we prompt LLMs to generate diverse outputs from three perspectives, \textit{Solution}, \textit{Specification} and \textit{Test case}, constructing a 3-partite graph. With two measure functions of consistency, we embed both inter- and intra-consistency information into the graph. The optimal choice of solutions is then determined based on analysis in the graph.
MPSC significantly boosts performance of foundation models (ChatGPT in this paper) on various benchmarks, including HumanEval (+\textbf{15.91}\%), MBPP (+\textbf{6.43}\%) and CodeContests (+\textbf{9.37}\%), even surpassing GPT-4.\footnote{\hspace{4pt}The code is available at \url{https://github.com/skpig/MPSC}.}
\end{abstract}

\section{Introduction}
In recent years, pre-trained large language models (LLMs) 
have demonstrated unprecedented proficiency in understanding, generating, and reasoning with human language \citep{brown2020language,chowdhery2022palm,openai2023gpt,touvron2023llama}. 
Among the diverse applications of LLMs, code generation stands out as pivotal task and has been acknowledged as a fundamental task for benchmarking \cite{liang2023holistic}. 
This task entails models to generate source codes from provided natural language intents. Many foundation models have exhibited remarkable zero-shot performance in code generation, such as ChatGPT and GPT4 \citep{openai2023gpt}, with successful deployments in real-world applications like Github Copilot. 

\begin{figure}[tb]
    \centering
    \includegraphics[width=0.45\textwidth]{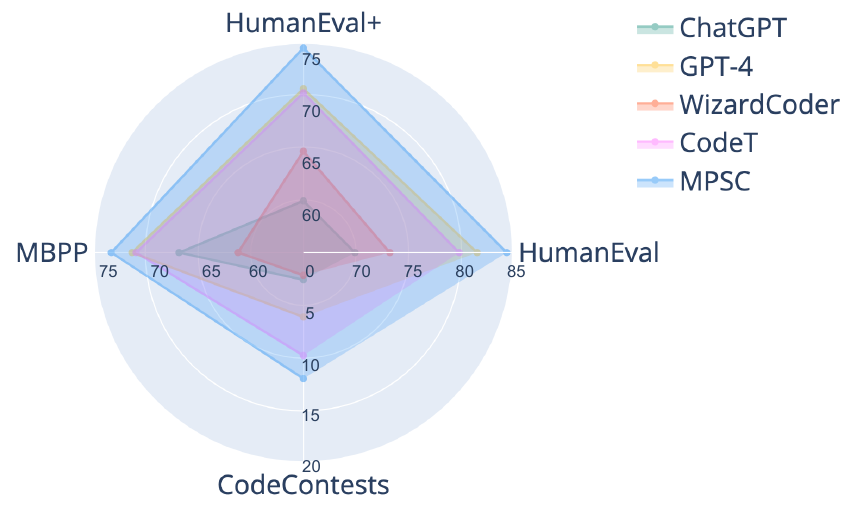}
    \caption{Pass$@1$ of MPSC. With ChatGPT as the foundation model, MPSC even surpasses GPT-4 and achieves SOTA performance on all four benchmarks.}
    \label{fig:intro_perf}
\end{figure}
Despite the remarkable abilities, LLMs often struggle to generate the correct code in a single attempt. 
Therefore, previous works sample diverse codes from LLMs and re-rank them by introducing \textit{verification properties} from software engineering.
For example, CodeT \cite{chen2022CodeT} generates test cases as a verification property, while ALGO \cite{zhang2023ALGO} generates oracles, the brute-force version of desired algorithms, as a verification property. These methods are essentially variants of majority voting, making the assumption that the correctness of experts (i.e. the verification properties) is better than that of choices (i.e. the desired code outputs). However, both verification properties and desired code outputs are usually generated by the identical model with respect to the same question, and hence the preference over verification properties is not always correct.


Instead, we believe that both desired code outputs and verification properties should be treated equally, since they are different \textit{perspectives} of LLM's deliberate thinking process in face of identical questions.
Aggregating various outputs from different perspectives can lead to a more credible result. To achieve this, we propose the \textbf{Multi-Perspective Self-Consistency (MPSC)} framework that incorporates both inter-consistency across multiple perspectives and intra-consistency within a single perspective. In this way, MPSC can fully leverage the consistency information within LLMs and select the model output with the most consistent functionality as the final answer.

In our framework, various verification techniques from software engineering can be included as extended perspectives for better reasoning. Specifically, we prompt the LLM to simultaneously generate diverse outputs from three well-established perspectives in software engineering, namely \textit{Solution}, \textit{Specification} and \textit{Test case} \cite{abrahamsson2017agile}. Solutions implement the desired functionality, specifications demonstrate the intended properties in formal language, while test cases outline the expected behavior for some specific inputs. 
Then, we treat these model outputs as vertices in a graph, and establish connections (i.e. edges) based on the pairwise agreement of vertices from different perspectives. 
Our goal is to identify the most reliable output using a score function, which evaluates all vertices by considering both intra- and inter-consistency information encoded in the graph. 
Specifically, the intra-consistency information guides the function to favor the most internally consistent output within a single perspective, while inter-consistency ensures that the scores for two outputs from different perspectives are similar if they reach a consensus.
We formalize the learning process of the score function as an optimization problem adhering to these two consistency criteria and leverage an iterative algorithm proposed by \citet{zhou2003Ranking} to achieve this goal.

We evaluate MPSC on four widely used code generation benchmarks, including HumanEval \citep{chen2021Evaluating}, HumanEval+ \citep{liu2023evalplus}, MBPP \citep{austin2021Program} and CodeContests \citep{li2022CompetitionLevel}. Experimental results show that our method boosts the performance of ChatGPT by a large margin, 15.91\% in HumanEval, 15.64\% in HumanEval+, 6.43\% in MBPP and 9.37\% in CodeContests. 
Our framework even surpasses GPT-4 \citep{openai2023gpt} as shown in Figure \ref{fig:intro_perf}. 


\begin{figure*}[t]
\begin{center}
\includegraphics[width=\linewidth]{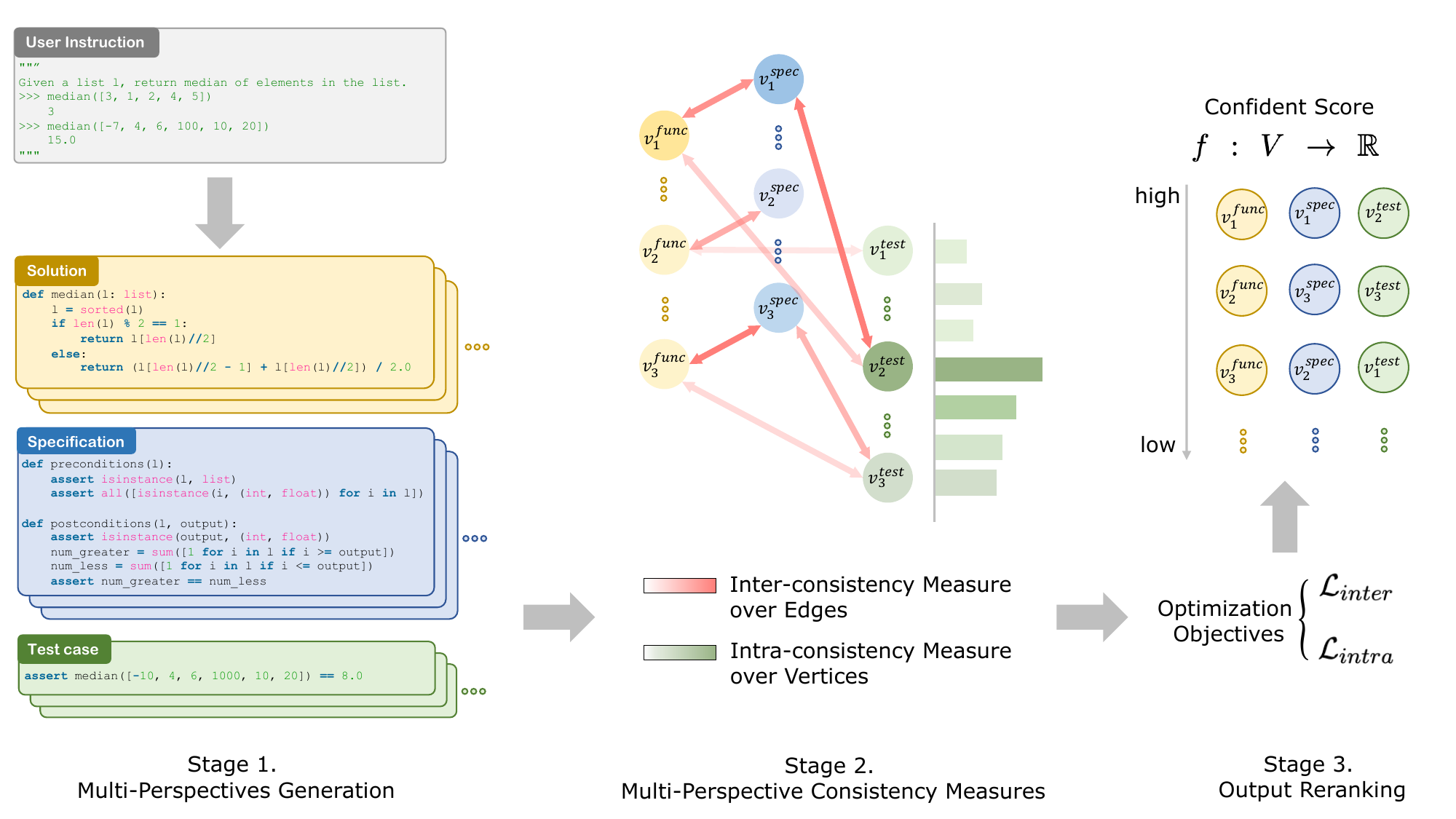 }
\end{center}
\caption{Overview of our MPSC code generation method. (a) Stage 1: we require a LLM to generate diverse solutions, specifications and test cases. A detailed example of the three perspectives of function \lstinline|median(l)| from HumanEval is presented. (b) Stage 2: we construct a 3-partite graph based on the generated outputs and then calculate both inter- and intra-consistency measures over edges and vertices respectively. The magnitudes of measurements are demonstrated by the shade of colors. (c) Stage 3: Incorporating the multi-perspective consistency information, we then learn a score function to re-rank outputs within each perspective.}
\label{fig:three_perspectives}
\end{figure*}
\section{Multi-Perspective Self-Consistency}

A single perspective can often lead to an incomplete understanding of a problem, akin to the parable of ``blind men describing an elephant". The reasoning process of LLMs follows a similar pattern. LLMs generally cannot guarantee the correctness of generated output in a single attempt, especially in code generation, which necessitates proficient natural language understanding, deliberate reasoning and rigorous controllability.

However, a key aspect of human intelligence is the ability to think from multiple perspectives, resulting in a more comprehensive understanding of situations and more accurate solutions to problems.
Inspired by the human cognition process, we propose a novel code generation method by reasoning from three well-established perspectives, solutions, specifications and test cases. Although noisy outputs may inevitably be included in the generated outputs of every perspective, we can leverage both intra-consistency and inter-consistency among the diverse outputs to distinguish the best ones from the noise. An overview of our proposed MPSC method is illustrated in Figure \ref{fig:three_perspectives}.

\vspace{-5pt}
\subsection{Solution, Specification and Test Case}

Given a user intent in natural language, we introduce solution, specification and test case as three perspectives to describe the desired functionality. A \textit{solution} is the source code implementing the functionality denoted as $g:\sX\rightarrow\sY$, which is also the target of code generation. A \textit{test case} is a pair of valid input and output satisfying the required functionality denoted as $(x,y)\in \sX\times\sY$. 
\textit{Specification} draws inspiration from \textit{Formal Verification} in software engineering, which mathematically proves the correctness of one program by ensuring its satisfaction of some certain formal specifications. In the context of software engineering, formal verification is usually written in formal programming languages, e.g. Coq \citep{coq} and Dafny \citep{dafny}, and conducted by accompanying verification tools. 
For the generalization of the proposed method in different program language scenarios, we adopt the idea of formal verification and limit the specifications within pre-conditions and post-conditions, which can be written as functions in the same programming language like solutions, without struggling with formal languages. 
Specifically, a pre-condition constrains the requirements that a valid input should satisfy, while a post-condition constrains the relationships that a pair of valid inputs and outputs should satisfy. We denote them as $h^{pre}: \sX\rightarrow\{False,True\}$ and $h^{post}:\sX\times\sY\rightarrow\{False,True\}$. Detailed examples of outputs are shown in Figure \ref{fig:three_perspectives}.

\subsection{Graph Construction}\label{sec:graph_con}

We require LLMs to generate diverse outputs from all three perspectives. 
We employ an 3-partite graph representation to capture the relationships among these generated outputs. 
Specifically, we represent the generated solutions $\{g_1,g_2,...,g_I\}$ with a vertex set $V^{func}$, the specification set $\{(h^{pre}_1,h^{post}_1),...,(h^{pre}_J,h^{post}_J)\}$ with $V^{spec}$, the test case set $\{(x_1, y_1),...,(x_K,y_K)\}$ with $V^{test}$, and hence construct a vertex set $V = V^{func}\cup V^{spec}\cup V^{test}$.
With edges connecting each pair of vertices from two distinct sets, we construct an undirected 3-partite graph $\gG=(V,E)$.
Our goal is to leverage the graph to encode the multi-perspective consistency information, and then learn a score function $f: V\rightarrow \R$ (also a vector $\vf,\,\evf_i=f(v_i)$) from it to choose the most reliable output among all.




\subsection{Inter-Consistency between Different Perspectives}


We distinguish between two kinds of consistency based on the perspectives involved. Intra-consistency is defined as the degree to which a given output aligns with others within the same perspective, following the original definition in \citet{wang2022SelfConsistency}. Conversely, inter-consistency is defined as the degree of consensus between a pair of outputs from two different perspectives. 

With the well-established definitions of these three perspectives in software engineering, each output implicitly describes a latent functionality regardless of whether it is a solution, a specification or a test case. Consequently, we define the inter-consistency between two vertices from different perspectives as the alignment of their latent functionalities. And the most appealing aspect is that we can quantify the alignments with a code interpreter in a deterministic manner\footnote{We provide the Python code snippets implementing the verification in Appendix \ref{app:inter_consistency_implementation}.}. We formalize the inter-consistency as a measure function $\omega(\cdot,\cdot
): V \times V \rightarrow \R$ (also the adjacency matrix $\mW$, where $\emW_{i,j} = \omega(v_i, v_j)$) to weight different edges in different ways as shown in Table \ref{tab:inter_consistency_expression}.

\begin{table}[h]
\begin{center}
\scalebox{0.8}{
\begin{tabular}{ll}
\toprule
\multicolumn{1}{c}{\bf Vertex Types}  &\multicolumn{1}{c}{\bf Expression of $\omega(v_i, v_j)$}
\\ 
\midrule
$v_i\in V^{func},\,v_j\in V^{spec}$ & $\E_{x\in\sX}[\1_{h^{pre}_j(x)\rightarrow h^{post}_j(x,g_i(x))}]$ \\
$v_i\in V^{func},\,v_j\in V^{test}$ & $\1_{g_i(x_j)=y_j}$ \\
$v_i\in V^{spec},\,v_j\in V^{test}$ & $\1_{h^{pre}_i(x_j)\wedge h^{post}_i(x_j,y_j)}$\\
otherwise & 0 \\
\bottomrule
\end{tabular}
}
\end{center}
\caption{Mathematical expressions of different inter-consistency measures $\omega(\cdot,\cdot)$.}
\label{tab:inter_consistency_expression}
\vspace{-15pt}
\end{table}


We then derive an optimization objective based on inter-consistency measurements, 
\begin{equation}
    \small
    \mathcal{L}_{inter} = \sum_{(v_i,v_j)\in E}\emW_{i,j}(f(v_i)-f(v_j))^2 
    =\vf^T \mL\vf
    \label{eq:def_inter}
\end{equation}

, where $\mL=\mD-\mW$ is the laplacian matrix of the graph $\gG$\footnote{In our experiment, we use the symmetric normalized Laplacian $\mL^{sym}=\mD^{-\frac12}\mL\mD^{-\frac12}$ for more robust performance.}. The loss function is the weighted sum of the local variation of each edge on the graph. An underlying assumption is that \textit{a pair of outputs exhibiting consistency are either both correct or both incorrect}. Therefore, the difference between scores of two connected vertices should be constrained by the penalty corresponding to the degree of consistency, i.e. edge weight.

\subsection{Intra-Consistency within the Same Perspective}

Following \citet{wang2022SelfConsistency}, we define the intra-consistency of one generated output as its similarity to others within the same perspective, which is denoted as a function $\varphi(\cdot): V \rightarrow \R$ (also a vector $\vy,\, \evy_i=\varphi(v_i)$).


\citet{wang2022SelfConsistency} limits the consistency to mere equalities in final answers, thereby lacking efficacy when applied to open-form tasks. In the scenario of code generation, we extend the scope of intra-consistency to lexical and semantic similarities.

\paragraph{Lexical intra-consistency by Bayes risk.} Minimum Bayes risk decoding \citep{kumar2004Minimum} selects the hypothesis $\rh\in\sH$ that minimizes the expected loss $R(\rh)=\E_{\ry\sim P(\ry)}[L(\ry,\rh)]$ over the distribution of label $\ry$. Because of the unavailability of $P(\ry)$, $P(\rh)$ is usually used as a proxy distribution in practice. Then the Bayes risk can be rewritten as $R(\rh)=\sum_{\rh'\in\sH}L(\rh',\rh)\cdot P(\rh')$, which is in fact measure the consistency of $\rh$ over the hypothesis space. Specifically, we utilize negative BLEU metrics \citep{papineni2002bleu} as the loss function $L$ aiming at lexical similarity and assume the hypothesis distribution is uniform, i.e.
\begin{equation*}
    \small
    \varphi(v_i) = C\cdot\sum_{v_j\in K(v_i)}\text{BLEU}(v_i,v_j)
\end{equation*}, where $C$ is the normalizing constant so that measures of outputs in one perspective sum up to $1$, $K(v_i)$ represents the other outputs within the same perspective as $v_i$.

\paragraph{Semantic intra-consistency by structural equivalence.}


In the realm of graph theory, two vertices are deemed structurally equivalent if they share identical weighted connections with the same third-party vertices. Utilizing this equivalence relation, we delineate $V^{func}$, $V^{spec}$, and $V^{test}$ into their respective structural equivalence classes. Noted that the weights assigned to edges reflect the alignments of latent functionalities associated with the vertices, and hence outputs within each equivalence class can be regarded as exhibiting consistent functional behavior. Therefore, we define the intra-consistency measure based on the structural equivalence classes within each perspective. Suppose $v_i$ belongs to the solution set $V^{func}$. The structural equivalence class of $v_i$ is denoted as $S(v_i)\subset V^{func}$, and neighbor sets of $v_i$ can be partitioned into two subsets $\{N_t(v_i)|t\in\{spec, test\}\}$ depending on the perspective they belong to. Overall, the lexical intra-consistency measure is defined as the multiplication of these three sets, i.e.

\begin{equation*}
    \small
    \varphi(v_i) = C\cdot |S(v_i)|\cdot\prod_t|N_t(v_i)|
\end{equation*}, where $C$ is the normalizing constant. The notation is similar when $v_i$ belongs to other perspectives.



Intra-consistency is in fact an estimate of the LLM's uncertainty \citep{kuhn2023semantic,xiong2023llms} and reflects the confidence of the model on one specific output. Therefore, we can utilize the intra-consistency information as a supervision signal by ensuring the closeness between the score function $f$ and the intra-consistency measure $\varphi$ with Mean Squared Error (MSE),
\begin{equation}
    \small
    \mathcal{L}_{intra}=\frac12 \sum_{v_i\in V} |f(v_i)-\varphi(v_i)|^2=\frac12 ||\vf-\vy||^2
    \label{eq:def_intra}
\end{equation}


\subsection{Optimization Formulation}

After all, following the criteria of inter- and intra-consistency, we can then formulate the learning process of $f$ as an optimization problem that combines both $\mathcal{L}_{inter}$ (Eq. \ref{eq:def_inter}) and $\mathcal{L}_{intra}$ (Eq. \ref{eq:def_intra}):
\begin{equation}
    \small
    \min_{f:V\rightarrow \R}\{\alpha\cdot\mathcal{L}_{inter}+(1-\alpha)\cdot\mathcal{L}_{intra}\}
\end{equation}

To solve this optimization problem on the graph, we adopt the iterative algorithm proposed by \citet{zhou2003Ranking}. The details of the algorithm can be found in Appendix \ref{app:alg_proof}.

\section{Experiment}

\subsection{Experiment Settings}

\paragraph{Dataset and metrics.} We conduct experiments on four widely used Python code generation benchmarks, including HumanEval, HumanEval+, MBPP and CodeContests. 
HumanEval \citep{chen2021Evaluating} and MBPP \citep{austin2021Program} are two hand-written Python programming problems. HumanEval+ \citep{liu2023evalplus} adds more unit tests based on HumanEval. CodeContests \citep{li2022CompetitionLevel} is a much more challenging dataset consisting of competition problems from the Codeforces platform. 
The evaluation metric is Pass$@k$ \citep{chen2021Evaluating}, which is
an unbiased estimator of the probability that at least one out of the $k$ solutions generated by the model passes unit tests. Details about the metric can be found in Appendix \ref{app:metrics}.

\paragraph{Implementation and baselines.} 
We compare several baselines from different LLMs for code like ChatGPT\footnote{https://chat.openai.com/} (i.e. GPT-3.5-Turbo), GPT-4 \citep{openai2023gpt}, Code Llama \citep{rozière2024code}, WizardCoder \citep{luo2023wizardcoder} and Deepseek Coder \citep{guo2024deepseekcoder}, to other post-hoc approaches enhancing LLMs during inference, including Self-consistency \citep{wang2022SelfConsistency}, \textsc{MBR-exec} \citep{shi2022Natural}, CodeT \citep{chen2022CodeT} and Self-collaboration \citep{dong2023Selfcollaboration}.
For both MPSC and other post-hoc augmentation approaches, we employ GPT-3.5-Turbo as the foundation model to generate 200 solutions. MPSC additionally generates 50 specifications and 100 test cases for each problem. Following the original setting in \citet{chen2022CodeT}, we additionally generate 500 test cases for other baselines.

\paragraph{Variants of MPSC.}
As shown in Eq.\ref{eq:def_intra}, the intra-consistency measure $\varphi(\cdot)$ is essentially used as a supervision signal without leveraging the semantics of ``consistency". Therefore, we include two variants in our experiments: (1) MPSC-Uniform is the baseline without any prior intra-consistency information and treats every vertex equally, i.e. $\varphi(v_i)=C$. (2) MPSC-Label includes the public example test cases in docstrings as silver labels, i.e. $\varphi(v_i)=C\cdot 1_{v_i\text{ is label}}$\footnote{The number of example test cases is two in average. Noted that MBPP doesn't provide test cases in docstrings.}. Further details regarding the implementation of our method and baselines are provided in Appendix \ref{app:settings}.

\begin{table*}[!tb]  
\centering  
\small
\begin{tabular}{lllllll}  
\toprule    
\midrule  
\multicolumn{1}{c}{Benchmark} & \multicolumn{3}{c}{\textbf{HumanEval}} &
\multicolumn{3}{c}{\textbf{HumanEval+}} \\
\midrule
\multicolumn{1}{c}{Metric} & Pass@1 & Pass@2 & Pass@5 & Pass@1 & Pass@2 & Pass@5 \\   
\midrule  
GPT4 & 81.48 & \underline{86.31} & \textbf{90.46} & 70.52 & \underline{75.48} & \textbf{79.54} \\
GPT-3.5-Turbo & 68.38 & 76.24 & 83.15 & 58.75 & 66.58 & 73.96 \\
DeepSeekCoder & 79.30 & - & - & - & - & - \\
WizardCoder & 73.20 & - & - & -& - & - \\
Code Llama  & 62.20 & - & - & - & - & - \\
\midrule
Self-consistency & 73.86\textsubscript{+5.48} & 73.93\textsubscript{-2.31} & 74.10\textsubscript{-9.05} & 63.50\textsubscript{+4.75} & 64.70\textsubscript{-1.88} & 65.67\textsubscript{-8.29} \\
\textsc{MBR-exec} & 72.96\textsubscript{+4.58} & 76.47\textsubscript{+0.23} & 79.00\textsubscript{-4.15} & 62.12\textsubscript{+3.37} & 67.08\textsubscript{+0.50} & 71.38\textsubscript{-2.58} \\
CodeT & 78.05\textsubscript{+9.67} & 78.05\textsubscript{+1.81} & 78.30\textsubscript{-4.85} & 67.87\textsubscript{+9.12} & 68.75\textsubscript{+2.17} & 69.65\textsubscript{-4.31} \\
Self-collaboration & 74.40\textsubscript{+6.02} & - & - & - & - & - \\
\midrule
MPSC-Uniform & 74.17\textsubscript{+5.79} & 77.02\textsubscript{+0.78} & 78.53\textsubscript{-4.62} & 65.05\textsubscript{+6.30} & 69.76\textsubscript{+3.18} & 71.72\textsubscript{-2.24} \\
MPSC-Lexical & 82.32\textsubscript{+13.94} & 84.76\textsubscript{+8.52} & 86.48\textsubscript{+3.33} & \textbf{74.39\textsubscript{+15.64}} & 75.00\textsubscript{+8.42} & 77.24\textsubscript{+3.28} \\
MPSC-Semantic & \underline{83.38\textsubscript{+15.00}} & 84.25\textsubscript{+8.01} & 84.45\textsubscript{+1.30} & \underline{73.54}\textsubscript{+14.79} & 74.46\textsubscript{+7.88} & 75.26\textsubscript{+1.30} \\
MPSC-Label & \textbf{84.29\textsubscript{+15.91}} & \textbf{86.79\textsubscript{+10.55}} & \underline{87.13\textsubscript{+3.98}} & 73.47\textsubscript{+14.72} & \textbf{76.66\textsubscript{+10.08}} & \underline{77.25\textsubscript{+3.29}} \\
\midrule  
\midrule  
\multicolumn{1}{c}{Benchmark} & \multicolumn{3}{c}{\textbf{MBPP}} &
\multicolumn{3}{c}{\textbf{CodeContests}} \\
\midrule
\multicolumn{1}{c}{Metric} & Pass@1 & Pass@2 & Pass@5 & Pass@1 & Pass@2 & Pass@5 \\   
\midrule  

GPT-4 & 71.26 & \textbf{74.27} & \textbf{76.99} & 6.1 & 8.28 & 11.72 \\
GPT-3.5-Turbo & 66.80 & 72.34 & \underline{76.60} & 2.57 & 4.22 & 7.16 \\  
DeepseekCoder & 70.00 & - & - & - & - & - \\
WizardCoder & 61.20 & - & - & 2.15 & 3.40 & 5.37 \\
Code Llama & 61.20 & - & - & - & - & - \\
\midrule
Self-consistency & 71.70\textsubscript{+4.90} & 71.73\textsubscript{-0.61} & 71.82\textsubscript{-4.78} & 8.10\textsubscript{+5.53} & 8.42\textsubscript{+4.20} & 8.48\textsubscript{+1.32} \\
\textsc{MBR-exec} & 70.79\textsubscript{+3.99} & 73.14\textsubscript{+0.80} & 74.85\textsubscript{-1.75} & 8.25\textsubscript{+5.68} & 8.87\textsubscript{+4.65} & 9.08\textsubscript{+1.92} \\
CodeT & \underline{71.90\textsubscript{+5.10}} & 71.95\textsubscript{-0.39} & 72.02\textsubscript{-4.58} & 9.92\textsubscript{+7.35} & 10.18\textsubscript{+5.96} & \underline{10.30}\textsubscript{+3.14} \\
Self-collaboration & 68.20\textsubscript{+1.40} & - & - & - & - & - \\
\midrule
MPSC-Uniform & 69.34\textsubscript{+2.54} & 70.06\textsubscript{-2.28} & 71.85\textsubscript{-4.75} & 4.71\textsubscript{+2.14} & 6.65\textsubscript{+2.43} & 8.31\textsubscript{+1.15} \\
MPSC-Lexical & 68.38\textsubscript{+1.58} & 70.26\textsubscript{-2.08} & 71.43\textsubscript{-5.17} & 5.45\textsubscript{+2.88} & 5.45\textsubscript{+1.23} & 6.06\textsubscript{-1.10} \\
MPSC-Semantic & \textbf{73.23\textsubscript{+6.43}} & \underline{73.29\textsubscript{+0.95}} & 73.55\textsubscript{-3.05} & \underline{10.09\textsubscript{+7.52}} & \underline{10.29\textsubscript{+6.07}} & \underline{10.30\textsubscript{+3.14}} \\
MPSC-Label & - & - & - & \textbf{11.94\textsubscript{+9.37}} & \textbf{15.55\textsubscript{+11.33}} & \textbf{18.20\textsubscript{+11.04}} \\
\midrule  
\bottomrule  
\end{tabular}  
\caption{The results on four code generation benchmarks. The foundation model for MPSC, Self-consistency, \textsc{MBR-exec}, CodeT, Self-collaboration are all GPT-3.5-Turbo. The improvements are calculated between methods and GPT-3.5-Turbo. The best and second best performance for each dataset are shown in \textbf{bold} and \underline{underline}. 
}  
\label{tab:main_result_new}  
\vspace{-5pt}
\end{table*}

\subsection{Main Results}

The experimental results on the four benchmarks are presented in Table \ref{tab:main_result_new}. We observe that MPSC consistently enhances the code generation capabilities of the foundation model (i.e. GPT-3.5-Turbo) across all benchmarks with a remarkable margin of improvement. Particularly, when $k$ is set to $1$, which is the most prevalent scenario in real-world applications, the performance improvement is notably significant (+15.91\% on HumanEval, +15.64\% on HumanEval+, +6.43\% on MBPP and +9.37\% on CodeContests). 
With the foundation model GPT-3.5-Turbo, our MPSC can even outperform GPT-4 in Pass@1 across all benchmarks.
Compared to other post-hoc augmentation approaches, even though they utilize more test cases, our MPSC still shows consistent advantages in all benchmarks, excluding the Pass@5 score in MBPP benchmark.
MPSC-Uniform serves as the bottom line of MPSC framework and still achieves great gains for the foundation model, which demonstrates that relying on inter-consistency proves to be entirely effective.
Moreover, incorporating various types of intra-consistency information leads to further improvements. Specifically, MPSC-Label and MPSC-Semantic exhibit particularly strong results. They are two representative approaches leveraging the external supervision signals or the internal consistency information respectively. Surprisingly, MPSC-Semantic can match or even surpass MPSC-Label in some benchmarks, which further highlights the significance of consistency information in LLMs. 
Besides, we also note that the performance of MPSC-Semantic and MPSC-Lexical remains largely unchanged as $k$ increases. This phenomenon aligns with the nature of MPSC, which assesses solutions based on their consistency within the foundation model. It implies that top-ranked solutions exhibit semantic similarity and are consistently either correct or incorrect. 
This reaffirms the capability of our proposed MPSC to effectively aggregate consistency information within LLMs, thereby selecting the most consistent answers. We have a more detailed discussion about this phenomenon in Appendix \ref{app:enhance_pass_at_5}.

\begin{table*}[htb]
    \centering
    \small
\begin{tabular}{lllllll}
\toprule
\midrule
\multicolumn{1}{c}{Benchmark} & \multicolumn{3}{c}{\textbf{HumanEval}} &
\multicolumn{3}{c}{\textbf{HumanEval+}} \\
\midrule
\multicolumn{1}{c}{Metric} & Pass@1 & Pass@2 & Pass@5 & Pass@1 & Pass@2 & Pass@5 \\   
\midrule
GPT-4 (gpt4-0614)~~~~ & 81.48 & 86.31 & 90.46 & 70.52 & 75.48 & 79.54\\
\multicolumn{1}{r}{+MPSC} & \textbf{92.15\textsubscript{+10.67}} & \textbf{91.62\textsubscript{+5.31}} & \textbf{91.80\textsubscript{+1.34}} & \textbf{81.72\textsubscript{+11.2}} & \textbf{81.77\textsubscript{+6.29}} & \textbf{82.12\textsubscript{+2.58}} \\
\midrule
WizzardCoder-34B$^\dag$ & 67.84 & 72.12 & \textbf{75.98} & 58.70 & 62.88 & \textbf{66.88} \\
\multicolumn{1}{r}{+MPSC} & \textbf{74.06\textsubscript{+6.22}} & \textbf{75.00\textsubscript{+2.88}} & 75.07\textsubscript{-0.91} & \textbf{65.45\textsubscript{+6.75}} & \textbf{65.71\textsubscript{+2.83}} & 66.19\textsubscript{-0.69} \\
\midrule
Code Llama-34B & 51.78 & 59.24 & 67.07 & 41.49 & 48.30 & 55.93 \\
\multicolumn{1}{r}{+MPSC} & \textbf{70.97\textsubscript{+19.19}} & \textbf{70.55\textsubscript{+11.31}} & \textbf{71.36\textsubscript{+4.29}} & \textbf{58.44\textsubscript{+16.95}} & \textbf{59.00\textsubscript{+10.70}} & \textbf{60.02\textsubscript{+4.09}} \\
\midrule
WizzardCoder-13B & 60.35 & 66.10 & 72.01 & 50.25 & 56.00 & 61.98 \\
\multicolumn{1}{r}{+MPSC} & \textbf{73.60\textsubscript{+13.25}} & \textbf{74.96\textsubscript{+8.86}} & \textbf{74.57\textsubscript{+2.56}} & \textbf{61.33\textsubscript{+11.08}} & \textbf{62.99\textsubscript{+6.99}} & \textbf{62.67\textsubscript{+0.69}} \\
\midrule
Code Llama-13B & 44.63 & 50.99 & 57.86 & 35.93 & 41.71 & 48.19 \\
\multicolumn{1}{r}{+MPSC} & \textbf{62.94\textsubscript{+18.31}} & \textbf{64.93\textsubscript{+13.94}} & \textbf{64.66\textsubscript{+6.80}} & \textbf{50.04\textsubscript{+14.11}} & \textbf{51.24\textsubscript{+9.53}} & \textbf{51.36\textsubscript{+3.17}} \\
\midrule
WizzardCoder-7B & 53.81 & 59.62 & 66.06 & 45.06 & 50.83 & 57.69\\
\multicolumn{1}{r}{+MPSC} & \textbf{63.85\textsubscript{+10.04}} & \textbf{64.04\textsubscript{+4.42}} & \textbf{67.32\textsubscript{+1.26}} & \textbf{53.69\textsubscript{+8.63}} & \textbf{55.07\textsubscript{+4.24}} & \textbf{59.45\textsubscript{+1.76}}\\
\midrule
Code Llama-7B & 39.38 & 45.18 & 52.79 & 34.33 & 39.18 & 45.25\\
\multicolumn{1}{r}{+MPSC} & \textbf{58.54\textsubscript{+19.16}} & \textbf{57.83\textsubscript{+12.65}} & \textbf{59.31\textsubscript{+6.52}} & \textbf{49.04\textsubscript{+14.71}} & \textbf{49.96\textsubscript{+10.78}} & \textbf{50.46\textsubscript{+5.21}} \\
\midrule
Deepseek Coder-6.7B & 71.73 & 80.92 & \textbf{86.73} & 61.72 & 71.42 & \textbf{78.54} \\
\multicolumn{1}{r}{+MPSC} & \textbf{82.38\textsubscript{+10.65}} & \textbf{83.92\textsubscript{+3.00}} & 84.71\textsubscript{-2.02} & \textbf{70.04\textsubscript{+8.32}} & \textbf{72.12\textsubscript{+0.70}} & 73.96\textsubscript{-4.58}\\
\midrule
\bottomrule
\end{tabular}
\caption{The performance of MPSC-Semantic with different foundation models. $^\dag$: We use nucleus sampling with temperature as 0.2 instead of greedy generation in this experiment. The best performance is shown in \textbf{bold}.}
\label{tab:generalization}
\vspace{-5pt}
\end{table*}

\subsection{Further Analysis}

\paragraph{Ablation study.} We conduct an ablation study to examine the impact of different perspectives on MPSC. The results are presented in Table \ref{tab:ablation}. Evidently, both the specification and test case perspectives play crucial roles in our framework. Additionally, test cases contribute more to the improvements than specifications. We attribute the observation to the better quality of test cases, as generating an accurate test case is considerably simpler than abstracting a comprehensive and sound specification.


\begin{table}[htb]
\centering
\small
\resizebox{\linewidth}{!}{
\begin{tabular}{lcccc}
\toprule
\multicolumn{1}{c}{Benchmark} & HumanEval & HumanEval+ & MBPP & CodeContests\\
\midrule
Ours & 83.38 & 73.54 & 73.23 & 10.09 \\
\quad w/o Specification & 82.32 & 73.52 & 70.18 & 9.17\\
\quad w/o Test case & 78.30 & 68.49 & 72.00 & 8.71\\
\quad w/o Both & 68.38 & 58.75 & 66.80 & 2.57\\
\bottomrule
\end{tabular}
}
\caption{The ablation study results (Pass$@1$) on four benchmarks.}
\label{tab:ablation}
\end{table}

\paragraph{Qualities of three perspectives.} We present the accuracy\footnote{Accuracy is equal to pass$@1$.} of generated solutions, specifications and test cases in Table \ref{tab:three_perspectives_acc}. 
The quality of all three perspectives is insufficient individually. Indeed, the generated verification properties (i.e. specifications and test cases) are even poorer than the generated solutions. It implies that prior works
\citep{zhang2023ALGO,chen2022CodeT}
relying on generated verification properties as experts for majority voting on solutions may fail, as these experts perform worse than the choices (i.e. solutions) themselves. Additionally, it indicates that the significant improvements brought by MPSC do not solely depend on the high quality of verification properties. The improvements come from the consistency information within LLMs, which helps to distinguish noise from high-quality solutions.

\begin{table}[htb]
\centering
\small
\begin{tabular}{cccc}
\toprule
\multicolumn{1}{c}{Perspective} & Solution & Specification & Test Case \\
\midrule 
HumanEval & 68.38 & 45.93 & 63.82\\
MBPP & 66.80 & 53.70  & 34.64\\

\bottomrule
\end{tabular}
\caption{Accuracy of solutions, specifications and test cases generated by GPT-3-Turbo.}
\label{tab:three_perspectives_acc}
\end{table}

\paragraph{Generalization over different LLMs.} 
MPSC is designed as a model-agnostic framework that assumes black-box access to the underlying foundation model. In assessing the extent of MPSC's generalization, we employ many other LLMs in addition to ChatGPT as foundation models. In specific, we consider the strongest model, GPT4, and three highly proficient open-source coding LLMs, Code Llama, WizardCoder and DeepSeek Coder in Python. The experimental results presented in Table \ref{tab:generalization} show that MPSC consistently yields significant improvements across all models, which demonstrates the robust generalization capabilities embedded in our proposed framework.

\paragraph{Impact of edge sparsity.}
\label{sec:edge_sparsity}

\begin{figure}[tbp]
    \centering
    \includegraphics[width=0.45\textwidth]{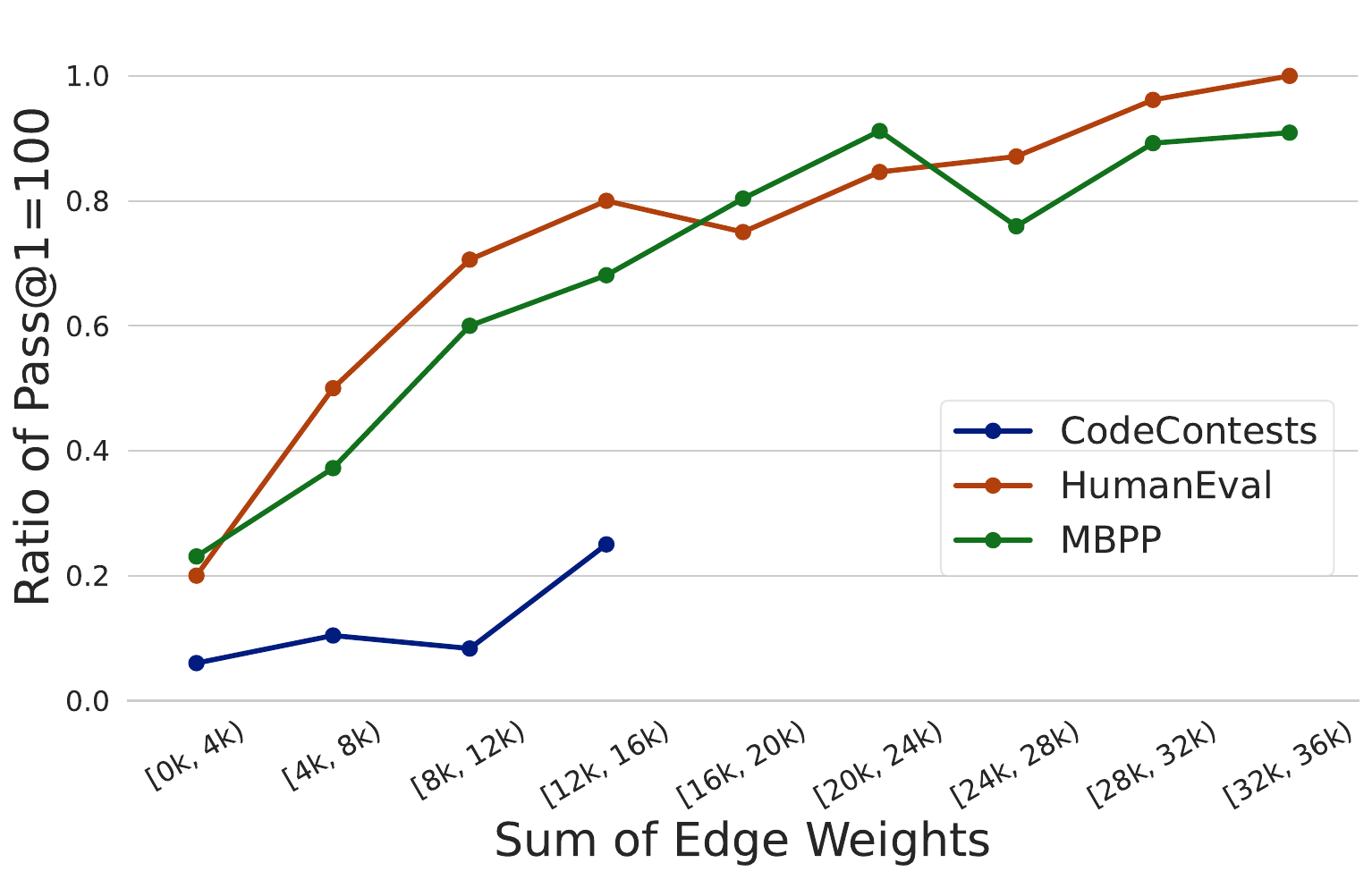}
    \captionof{figure}{The correlation between the performance of MPSC and the edge density.}
    \label{tab:edge_sparsity}
    \vspace{-10pt}
\end{figure}

Our framework significantly depends on the inter-consistency information between model outputs, which is represented as edges within the constructed graph. A critical question arises concerning the impact of edge sparsity on the framework's efficacy. To address this, we categorize all queries in the dataset into distinct bins based on the total edge weights in their corresponding graphs and compute the perfect performance ratio (i.e. Pass$@1$=100) for each bin. In this experiment, we employ the MPSC-Uniform configuration\footnote{The Uniform configuration is utilized to eliminate the influence of intra-consistency information.}. Figure \ref{tab:edge_sparsity} illustrates the correlation between edge density and performance. The results clearly demonstrate a positive correlation between the number of edges and the overall performance of our framework.

\paragraph{Impact of sampling number.} To examine the effect of the sampling number of different perspectives, we conduct an analysis experiment by varying the numbers of generated specifications and test cases. As shown in Table \ref{tab:sampling_humaneval}, MPSC constantly brings significant performance gains with varied specifications and test cases.
We note that MPSC generally suffers a slight degradation in performance when fewer specifications or test cases are used, which is consistent with our intuition. But the performance decline is relatively small (about 4\% with only 10\% of specifications and test cases retained). The observation indicts the remarkable performance and efficiency of MPSC, suggesting the potential for real-world application with reduced computational costs.

\begin{table}[tb]  
\small
\centering  
\resizebox{\linewidth}{!}{
\begin{tabular}{cccccc}  
\toprule
\multirow{2}{*}{\#Specification} & \multicolumn{5}{c}{\#Test Case}\\
\cmidrule{2-6}
~ & 10 & 20 & 50 & 100 & 200 \\
\midrule
10 & 78.93\textsubscript{+10.55} & 80.08\textsubscript{+11.7} & 83.82\textsubscript{+15.44} & 83.96\textsubscript{+15.58} & 84.17\textsubscript{+15.79}\\
20 & 77.17\textsubscript{+8.79} & 80.13\textsubscript{+11.75} & 83.82\textsubscript{+15.44} & 83.42\textsubscript{+15.04} & 85.44\textsubscript{+17.06}\\
50 & 80.23\textsubscript{+11.85} & 80.24\textsubscript{+11.86} & 82.24\textsubscript{+13.86} & 83.38\textsubscript{+15.00} & 83.57\textsubscript{+15.19}\\
100 & 80.39\textsubscript{+12.01} & 80.9\textsubscript{+12.52} & 80.92\textsubscript{+12.54} & 81.55\textsubscript{+13.17} & 84.17\textsubscript{+15.79}\\
\bottomrule
\end{tabular}
}
\caption{Pass$@1$ of MPSC-Semantic with different sampling numbers on HumanEval.}  
\label{tab:sampling_humaneval}
\end{table}

\section{Related Work}

\paragraph{Prompting techniques on consistency.}
Based on Chain-of-thought mechanism \citep{wei2022chain}, previous works have adopted various prompting techniques and decoding strategies to reveal the consistency of LLM outputs and further enhance the capabilities of LLMs. 
One line of approaches decodes multiple times from the same perspective and aggregate the results \citep{wang2022SelfConsistency,zhou2022least,jung2022maieutic,sun2022RecitationAugmented,chen2023universal}. For example, \citet{wang2022SelfConsistency} targets tasks with fixed answer sets and scores each answer based on the output frequency. Building on this, \citet{sun2022RecitationAugmented} introduces recitation as context for augmentation. 
While \citet{jung2022maieutic} focus on the two-value entailment relations (True or False) between statements and explanations. They treat the inference process as a weighted MAX-SAT problem and utilize a logistic solver to solve it. 
Another line draws inspiration from the ``Dual Process" in cognitive science, which posits that human reasoning is dominated by System 1 and System 2 \citep{daniel2017thinking,sloman1996empirical}. 
As a result, these approaches require LLMs to play different roles like generator (i.e. System 1) and verifier (i.e. System 2), and optimize the result iteratively by a conversational way \citep{madaan2023SelfRefine,shinn2023Reflexion,zhu2023Solving}. 
\citet{xiong2023Examining} also proposes the concept of "inter-consistency". Instead of referring to the consistency within the same LLM, they focus to tackle the inter-inconsistency problem between different models with a formal debate framework. 

\paragraph{LLM for code.}

LLMs pretrained on large-scale code data have demonstrated strong capabilities in the field of code generation, such as Codex \citep{chen2021Evaluating}, AlphaCode \citep{li2022CompetitionLevel}, CodeGen \citep{nijkamp2023codegen}, InCoder \citep{fried2022incoder}, StarCoder \citep{li2023starcoder}, Code Llama \citep{roziere2023codellama}, WizardCoder \citep{luo2023wizardcoder},
DeepSeekCoder \citep{guo2024deepseekcoder}. 
However, they remain unreliable, particularly in scenarios involving complex input-output mappings. Because of the low tolerance of compilers and operating systems for bugs, the instability makes LLMs hard to deploy into real-world applications. Several methods have been proposed to mitigate the phenomenon \citep{shi2022Natural,chen2022CodeT,zhang2022Coder,key2022Speak,ni2023lever,dong2023Selfcollaboration,olausson2023Demystifying,chen2023Teaching,zhang2023ALGO}.
The line of work with the most direct relevance to ours is to re-rank generated solutions in a post-hoc manner.
For example, 
\citet{shi2022Natural} matches the execution results of generated solutions for minimum Bayes risk selection.
\citet{zhang2022Coder} prompts another model as a reviewer to check whether generated programs satisfy the given language instruction by measuring $p(\text{instruction}|\text{program})$.
CodeT \citep{chen2022CodeT} additionally generates test cases to verify the generated solutions.
Similarly, ALGO \citep{zhang2023ALGO} additionally generates exhaustive search algorithms as oracle programs to generate high quality test cases for verification.

\paragraph{Ranking on graph.}
In our framework, the final stage is in fact a ranking problem in graph. 
There exists some renowned graph ranking algorithms like PageRank \citep{Page1998PageRank} and HITS \citep{HITS}.
While our approach is inspired by manifold ranking \citep{zhou2003Ranking}, which is built upon a regularization framework on discrete spaces (i.e. graphs in this scenario) \citep{zhou2003Learning,zhou2004Regularization,zhou2005Regularization}.




\section{Conclusion}
In this paper, we present a novel code generation method, Multi-Perspective Self-Consistency (MPSC), aimed at enhancing the performance of LLMs in complex code generation tasks where a single attempt may not suffice to ensure the accuracy of the output. Our proposed MPSC strategy capitalizes on both intra- and inter-consistency across three perspectives, solutions, specifications and test cases, to identify the most reliable answer.
We systematically validate the effectiveness of MPSC through comprehensive experiments conducted on four widely used code generation benchmarks. The evaluation results demonstrate that MPSC outperforms existing methods and achieves the state-of-the-art performance on all of them.

\section*{Limitations}
\paragraph{Evaluation in the wild.} Even though MPSC has shown remarkable performance on most widely used code generation benchmarks, its effectiveness in real-world scenarios remains largely unexplored. Existing code generation benchmarks often present simplified code generation tasks compared to the intricacies encountered in actual code developments, where user intents are harder to understand and the desired functionalities are more complex.
\paragraph{Generalization to other tasks.}
Since our proposed MPSC framework is designed to be model-agnostic and task-agnostic. We only conduct experiments in the code generation task in this paper. 
Actually, MPSC can be applied to other textual generation tasks like math problem solving and question answering. 
However, unlike code generation, where code interpreter can measure the agreement between outputs in a deterministic way, assessing the agreement between natural language outputs is non-trivial. A general task-agnostic inter-consistency measure is to solely rely on LLMs, whose evaluation ability for arbitrary textual inputs has been demonstrated recently. We leave it for future works to discuss.




\section*{Acknowledgments}
This work was supported by National Key R\&D Program of China (2021YFF0901502), Beijing Science and Technology Program (Z231100007423011) and Key Laboratory of Science, Technology and Standard in Press Industry (Key Laboratory of Intelligent Press Media Technology). We sincerely thank Xinyu Hu, Mingqi Gao, Xiao Pu and Xiang Chen for providing insightful advice about this work. We also appreciate the anonymous reviewers for their helpful comments.

\bibliography{custom,code,llm,ranking}
\bibliographystyle{acl_natbib}

\appendix
\newpage
\section{Details of the Iterative Algorithm}
\label{app:alg_proof}
\paragraph{Description of algorithm.}
The iterative algorithm is shown in Algorithm \ref{alg}.
\begin{algorithm}
\DontPrintSemicolon
\caption{Iterative Optimization}
\label{alg}
\KwIn{degree matrix $\mD=\text{diag}(d_1,...,d_N)$,
initialization score vector $\vy$, 
weighted adjacency matrix $\mW$, threshold $\epsilon$

}
\KwOut{optimal confidence score vector $\vf^*$}
\Begin{
$\vf^{(0)}\longleftarrow\vy$\;
$\mT\longleftarrow\mD^{-\frac12}\mW\mD^{-\frac12}$\;
$i\longleftarrow 0$\;
\Do{$||\vf^{(i)}-\vf^{(i-1)}||\leq\epsilon$}{
$\vf^{(i+1)}\longleftarrow \alpha\mT\vf^{(i)}+(1-\alpha)\vy$\;
$i\longleftarrow i+1$\;
}
$\vf^*\longleftarrow\vf^{(i)}$\;
\KwRet{$\vf^*$}\;
}
\end{algorithm}

\paragraph{Proof of Convergence}
We first expand the expression of $\vf^{(n)}$ according to the recursive formula
\begin{align*}
    \vf^{(n)} &= \alpha\mT\vf^{(n-1)} + (1-\alpha)\vy\\
    &= (\alpha\mT)^{n-1}\vf^{(0)} + (1-\alpha)\sum_{i=0}^{n-1}(\alpha\mT)^{i}\vy\\
\end{align*}
Notice that $\mT$ is similar to a stochastic matrix $\mW\mD^{-1}=\mD^{\frac12}(\mD^{-\frac12}\mW\mD^{-\frac12})\mD^{-\frac12}=\mD^{\frac12}\mT\mD^{-\frac12}$. Therefore the eigenvalues of $\alpha\mT$ are in $[-\alpha,\alpha]$. With $\alpha\in(0,1)$, we have
\begin{align*}
    \lim_{n\rightarrow\infty}(\alpha\mT)^n&=0\\
    \lim_{n\rightarrow\infty}\sum_{i=0}^{n-1}(\alpha\mT)^{i}&=(\mI-\alpha\mT)^{-1}
\end{align*}
Therefore
\begin{align*}
    \vf^* = \lim_{n\rightarrow\infty}f^{(n)} = (1-\alpha)(1-\alpha\mT)^{-1}\vy
\end{align*}

\paragraph{Proof of Equivalence}
Denote the optimization function as 
\begin{align*}
\mathcal{F} &= \alpha \vf^T(\mI - \mD^{-\frac12}\mW\mD^{\frac12})\vf + \frac{(1-\alpha)}2(\vf-\vy)^2\\
&= \alpha \vf^T(\mI - \mT)\vf + \frac{(1-\alpha)}2(\vf-\vy)^2
\end{align*}

Differentiate $\mathcal{F}$ with respect to $\vf$, we have
\begin{align*}
    \frac{\partial \mathcal{F}}{\partial \vf} &= \alpha (\mI-\mT)\vf + (1-\alpha)(\vf-\vy)
\end{align*}

Let the derivatives to 0, the solution $\vf' = (1-\alpha)(\mI-\alpha\mT)^{-1}\vy = \vf^*$. Therefore, the iterative algorithm is actually optimizing the objective function.

\paragraph{Results of closed-form solution.}
Despite the existence of a closed-form solution for the optimization problem, the required matrix inversion operation is computationally expensive. Conversely, the iterative algorithm exhibits rapid convergence and demonstrates strong empirical performance. Consequently, we employ the iterative algorithm in our experiments. Additionally, we provide the performance of the closed-form solution in Table \ref{tab:closed-form-solution}. Our results indicate that the iterative algorithm achieves a performance on par with that of the direct computation of the closed-form solution.

\begin{table}[hbt]  
\centering  
\scalebox{0.8}{  
\begin{tabular}{lccc}  
\toprule    
\midrule  
\multicolumn{1}{c}{Metric} & Pass@1 & Pass@2 & \multicolumn{1}{c}{Pass@5} \\   
\midrule  
\multicolumn{4}{c}{\textbf{HumanEval}} \\
\midrule
MPSC-Lexical & 82.32/82.32 & 84.76/84.76 & 86.48/86.59 \\
MPSC-Semantic & 83.38/83.46 & 84.25/84.15 & 84.45/84.45 \\
\midrule  
\multicolumn{4}{c}{\textbf{HumanEval+}} \\
\midrule
MPSC-Lexical & 74.39/74.39 & 75.00/75.00 & 77.24/77.34 \\
MPSC-Semantic & 73.54/73.97 & 74.46/75.04 & 75.26/75.96 \\
\midrule  
\multicolumn{4}{c}{\textbf{CodeContests}} \\
\midrule
MPSC-Lexical & 5.45/5.45 & 5.45/5.45 & 6.06/6.06 \\
MPSC-Semantic & 10.09/10.26 & 10.29/10.30 & 10.30/10.30 \\
\midrule  
\multicolumn{4}{c}{\textbf{MBPP}} \\
\midrule
MPSC-Lexical & 68.38/68.38 & 70.26/70.26 & 71.43/71.43  \\
MPSC-Semantic & 73.23/73.27 & 73.29/73.55 & 73.55/73.56 \\

\midrule  
\bottomrule  
\end{tabular}  
}  
\caption{Performance of MPSC optimized by the iterative algorithm or calculating closed-form solution directly. The results are presented in form of \lstinline|(iterative algorithm / closed-form solution)|.}  
\label{tab:closed-form-solution}  
\end{table} 

\newpage
\section{Implementation of Inter-Consistency}
\label{app:inter_consistency_implementation}
We present the code snippets measuring the inter-consistency between each pair of perspectives in Listing \ref{lst:1}, \ref{lst:2}, \ref{lst:3}. After execution with Python interpreter, the \lstinline{final_result} is acquired as $\omega(v_i,v_j)$.

\begin{lstlisting}[language=Python,caption={Inter-consistency between specifications and test cases.},label={lst:1}]
"""Generated specifications"""
# Pre-conditions
def preconditions(input):
...

# Post-conditions
def postconditions(input, output):
...

"""Generated test cases"""
test_case = {'input': ...,
             'output': ...}

"""Check inter-consistency"""
def check():
    pass_result = None
    try:
        preconditions(test_case['input'])
        postconditions(test_case['input'], test_case['output'])
        pass_result = True
    except Exception as e:
        pass_result = False
    return pass_result
global final_result
final_result = check()
\end{lstlisting}

\begin{lstlisting}[language=Python,caption={Inter-consistency between solutions and specifications.},label={lst:2}]
"""Generated solutions"""
def entry_point(input):
...

"""Generated specifications"""
# Pre-conditions
def preconditions(input):
...

# Post-conditions
def postconditions(input, output):
...

"""Generated casual inputs"""
casual_inputs = [...]

"""Check inter-consistency"""
def check():
    pass_result = []
    for ci in casual_inputs:
        try:
            output = entry_point(ci)
            postconditions(ci, output)
            pass_result.append(True)
        except Exception as e:
            pass_result.append(False)
    return sum(pass_result) / len(pass_result)
global final_result
final_result = check()
\end{lstlisting}

\begin{lstlisting}[language=Python,caption={Inter-consistency between solutions and test cases.},label={lst:3}]
"""Generated solutions"""
def entry_point(input):
...

"""Generated test cases"""
test_case = {'input': ...,
             'output': ...}

"""Check inter-consistency"""
def check():
    try:
        output = entry_point(test_case['input'])
        pass_result = (output == test_case['output'])
    except Exception as e:
        pass_result = False
    return pass_result
global final_result
final_result = check()
\end{lstlisting}

\newpage
\section{Discussion about Pass$@k$}
\label{app:metrics}

\vspace{-5pt}

In this section, we discuss the flaws of Pass$@k$ in \citet{chen2021Evaluating} and propose a variant for evaluating methods involved selection and filtering.

\citet{chen2021Evaluating} propose an unbiased estimator called Pass$@k$, which estimates the probability of a model passing unit tests within $k$ attempts. In specific, \citet{chen2021Evaluating} first samples a total of $n$ solutions with $c$ of them are correct, randomly samples $k$ solutions for testing, and use the probability of passing tests for estimation,

\vspace{-10pt}
$$
\text{Pass}@k = 1-\frac{\binom{n-c}{k}}{\binom{n}{k}}
$$
\vspace{-5pt}

Although their implementation serves as an effective measure of the code generation ability of different foundation models (referred to as the first category of methods in the following), it is not suitable to evaluate methods involving filtering or selection during the inference stage \citep{li2022CompetitionLevel,chen2022CodeT} (referred to as the second category of methods in the following), as the $n$ generated solutions are identical.

To address the limitation, we implement a variant of Pass$@k$. We assume each method provides a score function over the $n$ generated solutions, which provides a unified view for the two method categories and hence enables a fair comparison. The first category can be regarded as assigning an identical score to each solution. Similar to the original definition of Pass$@k$, we evaluate the method by testing the top-$k$ solutions with the highest scores. As the score function may assign the same score to multiple solutions, the test result of the top-k  is not deterministic but an expected value.

Mathematically, let's assume that a method sequentially arranges outputs into an ordered list $\{s_1,...,s_n\}$, such that $ \forall i>j, s_i\preceq s_j$ according to their scores. We define a set of solutions $\sS^k=\{s_i| s_k\preceq s_i\}$, which represents the solution set selected by the method. Suppose the cardinalily of $\sS^k$ is $\hat n$, the number of correct solutions within $\sS^k$ is $\hat c$. Noted that $\hat n \geq k$, and thus we uniformly sample $k$ solutions $\{s'_1,...,s'_k\}$ from $\sS^k$ for estimation,
\vspace{-5pt}
\begin{align*}
    \text{Pass}@k\text{ (Ours)}&=\E_{s'_1,...,s'_k}[\1_{\cup_{i=1}^k s'_i\text{ is correct}}]\\
    &=\Pr(\cup_{i=1}^k s'_i\text{ is correct})\\
    &=1-\Pr(\cap_{i=1}^k s'_i\text{ is incorrect})\\
    &=1-\frac{\binom{\hat n-\hat c}{k}}{\binom{\hat n}{k}}
\end{align*}
\vspace{-10pt}

For a the first category of methods, $\hat n$ equals $n$ since it treats each solution equally. As a result, our implementation of Pass$@k$ is identical to the original implementation in \citet{chen2021Evaluating}. 


\paragraph{Pass$@k$ of MPSC when $k$ is large}
\label{app:enhance_pass_at_5}

As shown in Table \ref{tab:main_result_new}, the performance of MPSC-Semantic and MPSC-Lexical remains largely unchanged as $k$ increases. This phenomenon aligns with the nature of MPSC, which assesses solutions based on their consistency, hence assigning similar scores to solutions with similar semantics. Consequently, all solutions will aggregate into many clusters depending on whether the assigned scores are identical, resulting in a lack of diversity within each cluster.
Therefore, the pass rate of solutions selected by MPSC will remains constant even in more attempts (i.e. varying $k$), if the number of attempts is 
still less than the top-ranked cluster size. A trivial method to address the problem is to increase the diversity by adopting a round-robin selection from all clusters, rather than selecting solutions according to scores from highest to lowest. The corresponding results are presented in Table \ref{tab:old_metric}. 
We can see the performance of Pass$@5$ is improved compared with that of Table \ref{tab:main_result_new}.

\begin{table}[htb]  
\centering  
\small
\begin{tabular}{lccc}  
\toprule
\midrule
Method & Pass@1 & Pass@2 & Pass@5 \\ 
\midrule
\multicolumn{4}{c}{HumanEval} \\
\midrule
GPT-3.5-Turbo & 68.38 & 76.24 & 83.15\\ 
MPSC-Lexical & 82.32 & 84.76 & 86.59 \\ 
MPSC-Semantic &83.35 & 86.08 & 89.75 \\ 
\midrule
\multicolumn{4}{c}{HumanEval+} \\
\midrule
GPT-3.5-Turbo & 58.75 & 66.58 & 73.96   \\ 
MPSC-Lexical & 74.39 & 75.0 & 77.44\\ 
MPSC-Semantic &73.08 & 77.89 & 83.20\\ 
\midrule
\multicolumn{4}{c}{MBPP} \\
\midrule
GPT-3.5-Turbo & 66.80 & 72.34 & 76.60\\ 
MPSC-Lexical & 68.38 & 70.26 & 71.43 \\ 
MPSC-Semantic & 73.24 & 74.46 & 78.01 \\ 
\midrule
\multicolumn{4}{c}{CodeContests} \\
\midrule
GPT-3.5-Turbo & 2.57 & 4.22 & 7.16 \\ 
MPSC-Lexical & 5.45 & 5.45 & 6.06 \\ 
MPSC-Semantic & 10.07 & 10.18 & 11.39\\ 
\midrule
\bottomrule
\end{tabular}  

\captionof{table}{Performance of MPSC evaluated by the Pass$@k$ metric in a round-robin way.}
\label{tab:old_metric}
\end{table}

\newpage

\begin{table}[htb]  
    \centering  
    \small
    \begin{tabular}{l@{}>{\raggedleft\arraybackslash}lccc}  
    \toprule  
    \midrule  
    \multicolumn{2}{c}{Method} & Pass$@1$ & Pass$@2$ & Pass$@5$\\  
    \midrule  
    \multicolumn{5}{c}{HumanEval} \\
    \midrule
    MP&SC & 74.17 & 77.02 & 78.53 \\  
    & + 1 test case & 85.37 & 86.59 & 85.13\\  
    & + 2 test cases & 85.98 & 86.18 & 85.36 \\  
    & + 5 test cases & 88.41 & 89.23 & 88.69 \\  
    & + 10 test cases & 89.02 & 90.24 & 88.81\\  
    \midrule  
    \multicolumn{5}{c}{HumanEval+} \\
    \midrule
    MP&SC & 65.05 & 69.76 & 71.72 \\  
    & + 1 test case & 85.37 & 86.59 & 85.13\\  
    & + 2 test cases & 85.98 & 86.18 & 85.36 \\  
    & + 5 test cases & 88.41 & 89.23 & 88.69 \\  
    & + 10 test cases & 89.02 & 90.24 & 88.81\\  
    \midrule  
    \multicolumn{5}{c}{MBPP} \\
    \midrule
    MP&SC & 69.34 & 70.06 & 71.85 \\  
    & + 1 test case & 69.85 & 71.99 & 72.69 \\  
    & + 2 test cases & 70.78 & 72.46 & 73.04 \\  
    & + 5 test cases & 71.25 & 72.93 & 73.47 \\  
    & + 10 test cases & 71.72 & 73.4 & 73.27 \\  
    \midrule  
    \bottomrule  
    \end{tabular}  
    \caption{Performance of MPSC with different numbers of golden test cases.}
    \label{tab:golden_test_case}  
\end{table}  

\begin{table}[tb]
\centering
\small
\begin{tabular}{llll}
    \toprule
    \midrule
    \multicolumn{1}{c}{Metric} & Pass@1 & Pass@2 & Pass@5 \\   
    \midrule  
    \multicolumn{4}{c}{HumanEval} \\
    \midrule
    WizzardCoder-34B & 66.04\textsubscript{±1.27} & 70.95\textsubscript{±0.84} & 75.51\textsubscript{±0.43} \\
    \multicolumn{1}{r}{+MPSC} & 76.32\textsubscript{±1.82} & 76.50\textsubscript{±1.20} & 75.93\textsubscript{±0.68} \\
    \midrule
    WizzardCoder-13B & 58.62\textsubscript{±1.23} & 65.05\textsubscript{±0.75} & 71.81\textsubscript{±0.14}\\
    \multicolumn{1}{r}{+MPSC} & 74.96\textsubscript{±0.96} & 75.51\textsubscript{±0.44} & 75.33\textsubscript{±0.57} \\
    \midrule
    WizzardCoder-7B & 52.20\textsubscript{±2.17} & 58.16\textsubscript{±1.94} & 64.73\textsubscript{±1.68}\\
    \multicolumn{1}{r}{+MPSC} & 65.09\textsubscript{±1.83} & 65.54\textsubscript{±1.38} & 66.51\textsubscript{±0.83}\\
    \midrule
    \midrule  
    \multicolumn{4}{c}{HumanEval+} \\
    \midrule
    WizzardCoder-34B & 56.12\textsubscript{±1.83} & 60.54\textsubscript{±1.66} & 64.67\textsubscript{±1.56} \\
    \multicolumn{1}{r}{+MPSC} & 65.21\textsubscript{±0.20} & 64.85\textsubscript{±1.10} & 64.74\textsubscript{±1.36} \\
    \midrule
    WizzardCoder-13B & 49.42\textsubscript{±0.59} & 54.91\textsubscript{±0.77} & 60.81\textsubscript{±0.83}\\
    \multicolumn{1}{r}{+MPSC} & 63.12\textsubscript{±1.35} & 63.68\textsubscript{±0.63} & 63.77\textsubscript{±0.94}
 \\
    \midrule
    WizzardCoder-7B & 43.74\textsubscript{±2.34} & 49.48\textsubscript{±2.37} & 56.26\textsubscript{±2.30} \\
    \multicolumn{1}{r}{+MPSC} & 54.64\textsubscript{±1.21} & 56.55\textsubscript{±1.05} & 58.21\textsubscript{±0.88}
 \\
    \midrule
    \bottomrule
\end{tabular}
\caption{The average performance of MPSC with three sample sets under different random seeds. We use MPSC-Semantic configuration in this experiment.}
\label{tab:stability}
\end{table}

\subsection{Analysis of Other Perspectives}
MPSC not only selects the optimal output from the target perspective but also chooses outputs from auxiliary perspectives, thereby generating corresponding by-products. In the context of code generation, which is the primary focus of this paper, MPSC can additionally identify more reliable test cases and specifications. We evaluate the quality of these by-products and present the results in Table \ref{tab:other_perspective}. The experimental results demonstrate that MPSC is proficient in selecting high-quality outputs across all relevant perspectives.

\begin{table*}[htb]
\centering
\scalebox{0.8}{
\begin{tabular}{ccccccc}
\toprule
\midrule
Benchmark & \multicolumn{3}{c}{HumanEval} & \multicolumn{3}{c}{MBPP}\\
\midrule
Metric & Pass$@1$ & Pass$@2$ & Pass$@5$ & Pass$@1$ & Pass$@2$ & Pass$@5$ \\
\midrule
~ & \multicolumn{6}{c}{Specification} \\
\midrule
GPT-3.5-Turbo & 45.93&58.76&72.26 & 53.7&62.37&70.60\\
MPSC & 73.58&73.59&74.38 & 71.86&71.38&73.40\\
\midrule
~ & \multicolumn{6}{c}{Test case} \\
\midrule
GPT-3.5-Turbo & 63.83&80.71&93.23 & 34.64&44.32&53.19
 \\
MPSC & 96.95&96.95&96.95 & 55.72&55.95&57.18\\
\midrule
\bottomrule
\end{tabular}}
\caption{The quality of specifications and test cases selected by MPSC. They can also be evaluated in Pass$@k$. We use MPSC-Semantic configuration in this experiment.}
\label{tab:other_perspective}
\end{table*}

\begin{table*}[htb]
    \centering
    \small
\begin{tabular}{lllllll}
\toprule
\midrule
\multicolumn{1}{c}{Benchmark} & \multicolumn{3}{c}{\textbf{HumanEval}} &
\multicolumn{3}{c}{\textbf{HumanEval+}} \\
\midrule
\multicolumn{1}{c}{Metric} & Pass@1 & Pass@2 & Pass@5 & Pass@1 & Pass@2 & Pass@5 \\   
\midrule
WizzardCoder-34B$^\dag$ & 67.84 & 72.12 & \textbf{75.98} & 58.70 & 62.88 & \textbf{66.88} \\
\multicolumn{1}{r}{+CodeT} & 73.17\textsubscript{+5.33} & 73.17\textsubscript{+1.05} & 72.03\textsubscript{-3.95} & 64.21\textsubscript{+5.51} & 64.36\textsubscript{+1.48} & 63.41\textsubscript{-3.47} \\
\multicolumn{1}{r}{+MPSC} & \textbf{74.06\textsubscript{+6.22}} & \textbf{75.00\textsubscript{+2.88}} & 75.07\textsubscript{-0.91} & \textbf{65.45\textsubscript{+6.75}} & \textbf{65.71\textsubscript{+2.83}} & 66.19\textsubscript{-0.69} \\
\midrule
Code Llama-34B & 51.78 & 59.24 & 67.07 & 41.49 & 48.30 & 55.93 \\
\multicolumn{1}{r}{+CodeT} & 67.99\textsubscript{+16.21} & 68.17\textsubscript{+8.93} & 68.28\textsubscript{+1.21} & 55.26\textsubscript{+13.77} & 56.70\textsubscript{+8.4} & 57.90\textsubscript{+1.97} \\
\multicolumn{1}{r}{+MPSC} & \textbf{70.97\textsubscript{+19.19}} & \textbf{70.55\textsubscript{+11.31}} & \textbf{71.36\textsubscript{+4.29}} & \textbf{58.44\textsubscript{+16.95}} & \textbf{59.00\textsubscript{+10.70}} & \textbf{60.02\textsubscript{+4.09}} \\
\midrule
WizzardCoder-13B & 60.35 & 66.10 & 72.01 & 50.25 & 56.00 & 61.98 \\
\multicolumn{1}{r}{+CodeT} & 66.86\textsubscript{+6.51} & 67.18\textsubscript{+1.08} & 67.93\textsubscript{-4.08} & 58.23\textsubscript{+7.98} & 58.72\textsubscript{+2.72} & 58.99\textsubscript{-2.99} \\
\multicolumn{1}{r}{+MPSC} & \textbf{73.60\textsubscript{+13.25}} & \textbf{74.96\textsubscript{+8.86}} & \textbf{74.57\textsubscript{+2.56}} & \textbf{61.33\textsubscript{+11.08}} & \textbf{62.99\textsubscript{+6.99}} & \textbf{62.67\textsubscript{+0.69}} \\
\midrule
Code Llama-13B & 44.63 & 50.99 & 57.86 & 35.93 & 41.71 & 48.19 \\
\multicolumn{1}{r}{+CodeT} & 57.99\textsubscript{+13.36} & 58.25\textsubscript{+7.26} & 57.91\textsubscript{+0.05} & 50.03\textsubscript{+14.1} & 50.46\textsubscript{+8.75} & 50.48\textsubscript{+2.29} \\
\multicolumn{1}{r}{+MPSC} & \textbf{62.94\textsubscript{+18.31}} & \textbf{64.93\textsubscript{+13.94}} & \textbf{64.66\textsubscript{+6.80}} & \textbf{50.04\textsubscript{+14.11}} & \textbf{51.24\textsubscript{+9.53}} & \textbf{51.36\textsubscript{+3.17}} \\
\midrule
WizzardCoder-7B & 53.81 & 59.62 & 66.06 & 45.06 & 50.83 & 57.69\\
\multicolumn{1}{r}{+CodeT} & 63.17\textsubscript{+9.36} & 63.36\textsubscript{+3.74} & 63.41\textsubscript{-2.65} & \textbf{54.13\textsubscript{+9.07}} & 55.05\textsubscript{+4.22} & 55.74\textsubscript{-1.95} \\
\multicolumn{1}{r}{+MPSC} & \textbf{63.85\textsubscript{+10.04}} & \textbf{64.04\textsubscript{+4.42}} & \textbf{67.32\textsubscript{+1.26}} & 53.69\textsubscript{+8.63} & \textbf{55.07\textsubscript{+4.24}} & \textbf{59.45\textsubscript{+1.76}}\\
\midrule
Code Llama-7B & 39.38 & 45.18 & 52.79 & 34.33 & 39.18 & 45.25\\
\multicolumn{1}{r}{+CodeT} & 51.68\textsubscript{+12.30} & 51.83\textsubscript{+6.65} & 51.90\textsubscript{-0.89} & 44.06\textsubscript{+9.73} & 44.47\textsubscript{+5.29} & 44.71\textsubscript{-0.54} \\
\multicolumn{1}{r}{+MPSC} & \textbf{58.54\textsubscript{+19.16}} & \textbf{57.83\textsubscript{+12.65}} & \textbf{59.31\textsubscript{+6.52}} & \textbf{49.04\textsubscript{+14.71}} & \textbf{49.96\textsubscript{+10.78}} & \textbf{50.46\textsubscript{+5.21}} \\
\midrule
\bottomrule
\end{tabular}
\caption{The performance of MPSC-Semantic with different foundation models. $^\dag$: We use nucleus sampling with temperature as 0.2 instead of greedy generation in this experiment. The best performance is shown in \textbf{bold}.}
\label{tab:full_generalization}
\vspace{-5pt}
\end{table*}

\begin{table*}[htb]
\centering
\scalebox{0.7}{
\begin{tabular}{ccccccccccccc}
\toprule
\midrule
Benchmark & \multicolumn{3}{c}{HumanEval} & \multicolumn{3}{c}{HumanEval+} & \multicolumn{3}{c}{MBPP} & \multicolumn{3}{c}{CodeContest}\\
\midrule
Metric & Pass$@1$ & Pass$@2$ & Pass$@5$ & Pass$@1$ & Pass$@2$ & Pass$@5$ & Pass$@1$ & Pass$@2$ & Pass$@5$ & Pass$@1$ & Pass$@2$ & Pass$@5$ \\
\midrule
GPT-3.5-Turbo &	68.38 & 76.24 & 83.15 & 58.75 & 66.58 & 73.96 & 66.80 & 72.34 & 76.60 & 2.57 & 4.22 & 7.16\\
GPT-4 &	81.48 & 86.31 & 90.46 & 70.52 & 75.48 & 79.54 & 71.26 & 74.27 & 76.99 & 6.1 & 8.28 & 11.72\\
MPSC &	81.5 & 83.7 & 91.05 & 72.18 & 75.2 & 81.22 & 72.78 & 74.24 & 78.25 & 10.77 & 12.64 & 13.84\\
\midrule
\bottomrule
\end{tabular}}
\caption{The performance of MPSC with 100 solutions, 50 specifications and 50 test cases}
\label{tab:equal_cost}
\end{table*}

\section{More Analysis}
\label{app:other_analysis}


\subsection{MPSC with User-provided Test Cases} In practical applications of code generation, users often provide a limited number of test cases to outline the desired functionality, thereby assisting the model in generating code that aligns with the requirements. In Table \ref{tab:main_result_new}, MPSC-Label has shown remarkable performance. In this study, we investigate the potential performance improvements of the method in such scenarios by incorporating various quantities of golden test cases. These golden test cases are generated and then validated using canonical solutions provided in the benchmarks. We conduct experiments on the HumanEval and MBPP dataset \footnote{CodeContests doesn't provide canonical solutions. Therefore, we cannot conduct evaluation of test cases.} and present the results in Table \ref{tab:golden_test_case}. The substantial performance enhancements achieved with the inclusion of merely five golden test cases underscore the feasibility of implementing MPSC in user-interactive application scenarios.

\subsection{Time Overhead}

The additional time overhead imposed by MPSC mainly comes from inter-consistency measurements and the iterative algorithm (Alg. \ref{alg}). For the former, we need to process a total of $(200\times 50+50\times 100+100\times200)=35000$ edges with a time limit of $0.001$ seconds per edge. In contrast, CodeT \citep{chen2022CodeT}, the strongest baseline, requires to process a total of $200 \times 500=100000$ edges. Moreover, this process can be fully parallelized. For the latter, the iterative algorithm converges rapidly, typically within an average of about $50$ rounds, requiring less than $0.1$ second. Overall, the time overhead of MPSC is acceptable in most code development scenarios, laying the groundwork for real-world deployments.

\subsection{Stability of MPSC}

We explore the stability of MPSC with respect to the sampling process. We conduct the sampling process of WizardCoder with three random seeds and then assess the performance of MPSC on the generated solutions. The average results are shown in Table \ref{tab:stability}. It is evident that the improvement brought by MPSC is very stable.

\subsection{MPSC with Limited API Calls}
We here discuss another setting, where MPSC utilizes 100 solutions, 50 specifications and 50 test cases, requiring identical API calls to the foundation model baselines. The results shown in Table \ref{tab:equal_cost} again prove the supreme performance of MPSC over baselines under fair comparison.

\subsection{Comparison on Other Foundation Models}
Table \ref{tab:generalization} demonstrates the generalization of MPSC on other foundation models. We also conduct an experiment to compare MPSC with CodeT, the strongest baseline, under this setting. The results are presented in Table \ref{tab:full_generalization}.


\section{Experiment Settings and Baselines}
\label{app:settings}

We incorporate various baselines in code generation. First of all, we include many strong large language models like ChatGPT (gpt-3.5-turbo 0614 version), GPT-4 (gpt4-0614 version), Code Llama-Instruct-34B, WizardCoder-Python-34B and DeepSeekCoder-7B-Instruct. The specific hyper-parameters for inference of ChatGPT and GPT4 are shown in Table \ref{tab:hyper_llm}. MPSC requires an additional hyper-parameter $\alpha$, which controls the balance between inter-consistency and intra-consistency in the algorithm. Given that the quality of inter-consistency significantly depends on the edge density of the graph, we utilize the mean edge weight to determine the value of $\alpha$. Empirically, we assign a relatively small value of $\alpha$ (0.01) when the edges are sparse on the graph, indicated by the mean edge weight less than 0.16. Other, we assign a large value of $\alpha$ (0.95) otherwise to better leverage inter-consistency.

\begin{table}[htb]
    \centering
    \begin{tabular}{c|c}
    \toprule
    Temperature & 0.8 \\
    Top P  & 0.95 \\
    Frequency Penalty & 0\\
    Presence Penalty & 0\\
    \bottomrule
    \end{tabular}
    \caption{The Inference hyper-parameters of LLMs.}
    \label{tab:hyper_llm}
\end{table}

We also include several baselines like Self-Consistency \textsc{MBR-exec}, CodeT and Self-collaboration, which enhance the inference capability of LLMs in a post-hoc manner. 
\begin{itemize}[leftmargin=16pt]
    \item CodeT: This baseline first uses generated test cases to verify each solution by code execution. Then it utilizes RANSAC algorithm to create consensus sets based on execution results. The size of consensus set is then used to rank solutions. We generate 500 test cases for CodeT following the original implementation in \citet{chen2022CodeT}.
    \item Self-Consistency: We implement this baseline following \citet{chen2022CodeT}. If two solution pass the same set of generated test cases and specifications, we regard them “consistent”. Then we take a majority voting to rank solutions following \citet{wang2022SelfConsistency}.
    \item \textsc{MBR-exec}: This baseline ranks solutions by minimum Bayes risk decoding based on the execution results in the 500 generated test cases.
\end{itemize}

For a fair comparison between our proposed MPSC and these baselines, we employ the same solutions generated by ChatGPT for them to rerank. In specific, we sample 200 solutions following the conventional setting. Since some methods leverage generated test cases and specifications as well, we use the same set of test cases and specifications generated by ChatGPT for both MPSC and these baselines.


We don't include ALGO \cite{zhang2023ALGO} as baseline, because it requires to keep generating oracle programs until they pass all public example test cases, whose complexity is unlimited.

\newpage
\section{Prompt for MPSC}
\begin{tcolorbox}[title = {Prompt for Generating Specifications}]
I want you to act as a python programmer. Given a docstring about a python method, you need to write its pre-conditions in one test function ``def preconditions(input)" and post-conditions in another test function ``def postconditions(input, output):". You should ensure invalid input or output of the method will raise error in the two test functions.\\

\lstinline{```}Python\\
\{\textcolor{purple}{Demonstration Docstrings 1}\}\\
\qquad pass\\
\#Pre-conditions\\
\{\textcolor{purple}{Demonstration Preconditions 1}\}\\
\#Post-conditions\\
\{\textcolor{purple}{Demonstration Postconditions 1}\}\\
\lstinline{```}\\
\\
\lstinline{```}Python\\
\{\textcolor{purple}{Demonstration Docstrings 2}\}\\
\qquad pass\\
\#Pre-conditions\\
\{\textcolor{purple}{Demonstration Preconditions 2}\}\\
\#Post-conditions\\
\{\textcolor{purple}{Demonstration Postconditions 2}\}\\
\lstinline{```}\\
\\
\lstinline{```}Python\\
\{\textcolor{blue}{Docstrings}\}\\
\qquad pass\\

\end{tcolorbox}

\begin{tcolorbox}[title = {Prompt for Generating Solutions}, float, floatplacement=h!]
I want you to act like a Python programmer. I will give you the declaration of a function and comments about its property. You need to implement the body of the function in the code block. Do not modify any code I provide. Do not provide any explanations.\\
\\
Here is the question.\\
\lstinline{```}Python\\
\{\textcolor{blue}{Docstrings}\}\\
\lstinline{```}\\
\label{tab:prompt1}
\end{tcolorbox}

\begin{tcolorbox}[title = {Prompt for Generating Test Cases}]

\lstinline{```}Python\\
\# Given a docstring, continue to write the following code with 10 valid assertion statements to check the correctness of the function. Provide diverse test cases. \\
\{\textcolor{blue}{Docstrings}\}\\
\qquad pass\\
\\
\# check the correctness of with 10 different valid assertion statements in the form of ``assert \{\textcolor{purple}{entry point}\}(...) == ..."\\
assert \\\\
\label{tab:prompt2}
\end{tcolorbox}

\section{Case Study}
We present some cases from HumanEval, MBPP, CodeContests in the following pages.

\begin{lstlisting}[language=Python,caption={Cases from HumanEval.},float=*t,linewidth=\linewidth]
"""User Intent"""
def get_row(lst, x):
    """
    You are given a 2 dimensional data, as a nested lists,
    which is similar to matrix, however, unlike matrices,
    each row may contain a different number of columns.
    Given lst, and integer x, find integers x in the list,
    and return list of tuples, [(x1, y1), (x2, y2) ...] such that
    each tuple is a coordinate - (row, columns), starting with 0.
    Sort coordinates initially by rows in ascending order.
    Also, sort coordinates of the row by columns in descending order.
    
    Examples:
    get_row([
      [1,2,3,4,5,6],
      [1,2,3,4,1,6],
      [1,2,3,4,5,1]
    ], 1) == [(0, 0), (1, 4), (1, 0), (2, 5), (2, 0)]
    get_row([], 1) == []
    get_row([[], [1], [1, 2, 3]], 3) == [(2, 2)]
    """


"""Solution selected by MPSC"""
def get_row(lst, x):
    result = []
    for row_index, row in enumerate(lst):
        for col_index, val in reversed(list(enumerate(row))):
            if val == x:
                result.append((row_index, col_index))
    result.sort(key=lambda x: x[0])
    return result


"""Solution selected by CodeT"""
def get_row(lst, x):
    result = []
    for i, row in enumerate(lst):
        if x in row:
            for j, val in sorted(enumerate(row), key=lambda x: -x[1]):
                if val == x:
                    result.append((i, j))
    return result
\end{lstlisting}

\begin{lstlisting}[language=Python,caption={Cases from MBPP.},float=*t,linewidth=\linewidth]
"""User Intent"""
def count_first_elements(test_tup):
    '''
    Write a function to find the number of elements that occurs before the tuple element in the given tuple.
    '''


"""Solution selected by MPSC"""
def count_first_elements(test_tup):
    count = 0
    for i in range(len(test_tup)):
        if isinstance(test_tup[i], tuple):
            break
        count += 1
    return count


"""Solution selected by CodeT"""
def count_first_elements(test_tup):
    count = 0
    for i in range(len(test_tup)):
        if test_tup[i] == test_tup[0]:
            break
        count += 1
    return count




\end{lstlisting}


\begin{lstlisting}[language=Python,float=*t,linewidth=\linewidth]
"""User Intent"""
def solution(inputs: str) -> str:
    '''
    You are given an array a of length n.
    Let's define the eversion operation. Let x = a_n. Then array a is partitioned into two parts: left and right. The left part contains the elements of a that are not greater than x (<= x). The right part contains the elements of a that are strictly greater than x (> x). The order of elements in each part is kept the same as before the operation, i. e. the partition is stable. Then the array is replaced with the concatenation of the left and the right parts.
    For example, if the array a is [2, 4, 1, 5, 3], the eversion goes like this: [2, 4, 1, 5, 3] -> [2, 1, 3], [4, 5] -> [2, 1, 3, 4, 5].
    We start with the array a and perform eversions on this array. We can prove that after several eversions the array a stops changing. Output the minimum number k such that the array stops changing after k eversions.

    Input:
    Each test contains multiple test cases. The first line contains the number of test cases t (1 <= t <= 100). Description of the test cases follows.
    The first line contains a single integer n (1 <= n <= 2 * 10^5).
    The second line contains n integers a_1, a_2, ..., a_n (1 <= a_i <= 10^9).
    It is guaranteed that the sum of n over all test cases does not exceed 2 * 10^5.

    Output:
    For each test case print a single integer k - the number of eversions after which the array stops changing.
    Example
    assert solution('3\n5\n2 4 1 5 3\n5\n5 3 2 4 1\n4\n1 1 1 1') == '1\n2\n0'
    '''
"""Solution selected by MPSC"""
def solution(inputs: str) -> str:
    def eversion(a):
        n = len(a)
        l, r = [], []
        x = a[-1]
        for i in range(n - 1):
            if a[i] <= x:
                l.append(a[i])
            else:
                r.append(a[i])
        return l + [x] + r

    def is_same(a, b):
        return len(a) == len(b) and all(ai == bi for ai, bi in zip(a, b))

    res = []
    inputs = inputs.strip().split('\n')
    t = int(inputs[0])
    idx = 1
    for _ in range(t):
        n = int(inputs[idx])
        a = list(map(int, inputs[idx+1].split()))
        idx += 2
        cnt = 0
        while True:
            new_a = eversion(a)
            cnt += 1
            if is_same(a, new_a):
                break
            a = new_a
        res.append(str(cnt - 1))
    return '\n'.join(res)

\end{lstlisting}
\begin{lstlisting}[language=Python,caption={Cases from CodeContests.},float=*t,linewidth=\linewidth,firstnumber=53]
"""Solution selected by CodeT"""
def solution(inputs: str) -> str:
    # split the input into list of test cases
    test_cases = inputs.strip().split('\n')[1:]
    result = []
    
    # loop through each test case
    for i in range(0, len(test_cases), 2):
        n = int(test_cases[i])
        a = list(map(int, test_cases[i+1].split()))
        k = 0
        
        # perform eversion operation until array stops changing
        while True:
            x = a[-1]
            left, right = [], []
            for j in range(n-1):
                if a[j] <= x:
                    left.append(a[j])
                else:
                    right.append(a[j])
            new_a = left + [x] + right
            if new_a == a:
                break
            a = new_a
            k += 1
        
        result.append(str(k))
    
    # join the results and return
    return '\n'.join(result)


\end{lstlisting}

\end{document}